\documentclass[letterpaper]{article} 
\usepackage{aaai2026}  
\nocopyright
\usepackage{times}  
\usepackage{helvet}  
\usepackage{courier}  
\usepackage[hyphens]{url}  
\usepackage{graphicx} 
\usepackage{natbib}  
\usepackage{caption} 
\usepackage{algorithm}
\usepackage{algorithmic}

\usepackage{newfloat}
\usepackage{listings}
\DeclareCaptionStyle{ruled}{labelfont=normalfont,labelsep=colon,strut=off} 
\floatstyle{ruled}
\newfloat{listing}{tb}{lst}{}
\floatname{listing}{Listing}

\usepackage{booktabs}
\usepackage{subcaption}
\usepackage{amsfonts}

\usepackage[section]{placeins}

\title{Aggregating Local Saliency Maps for Semi-Global Explainable Image Classification}
 \author{
     James Hinns,
     David Martens
 }
 \affiliations{
     University of Antwerp, Belgium
 }

\begin{document}

\maketitle

\begin{abstract}
Deep learning dominates image classification tasks, yet understanding how models 
arrive at predictions remains a challenge. Much research focuses on local explanations 
of individual predictions, such as saliency maps, which visualise the influence of
specific pixels on a model's prediction. 
However, reviewing many of these explanations to identify recurring patterns is infeasible, 
while global methods often oversimplify and miss important local behaviours.
To address this, we propose Segment Attribution Tables (SATs), a method for summarising 
local saliency explanations into \mbox{(semi-)global} insights. SATs take image segments (such as 
``eyes'' in Chihuahuas) and leverage saliency maps to quantify their influence. These segments highlight 
concepts the model relies on across instances and reveal spurious correlations, such as 
reliance on backgrounds or watermarks, even when out-of-distribution test performance sees little change.
SATs can explain any classifier for which a form of saliency map can be produced, using segmentation maps that provide named segments.
SATs bridge the gap between oversimplified global summaries and 
overly detailed local explanations, offering a practical tool 
for analysing and debugging image classifiers.
\end{abstract}

\vspace{-1em}
\section{Introduction}

\begin{figure*}[t]
  \centering
  \includegraphics[width=\textwidth]{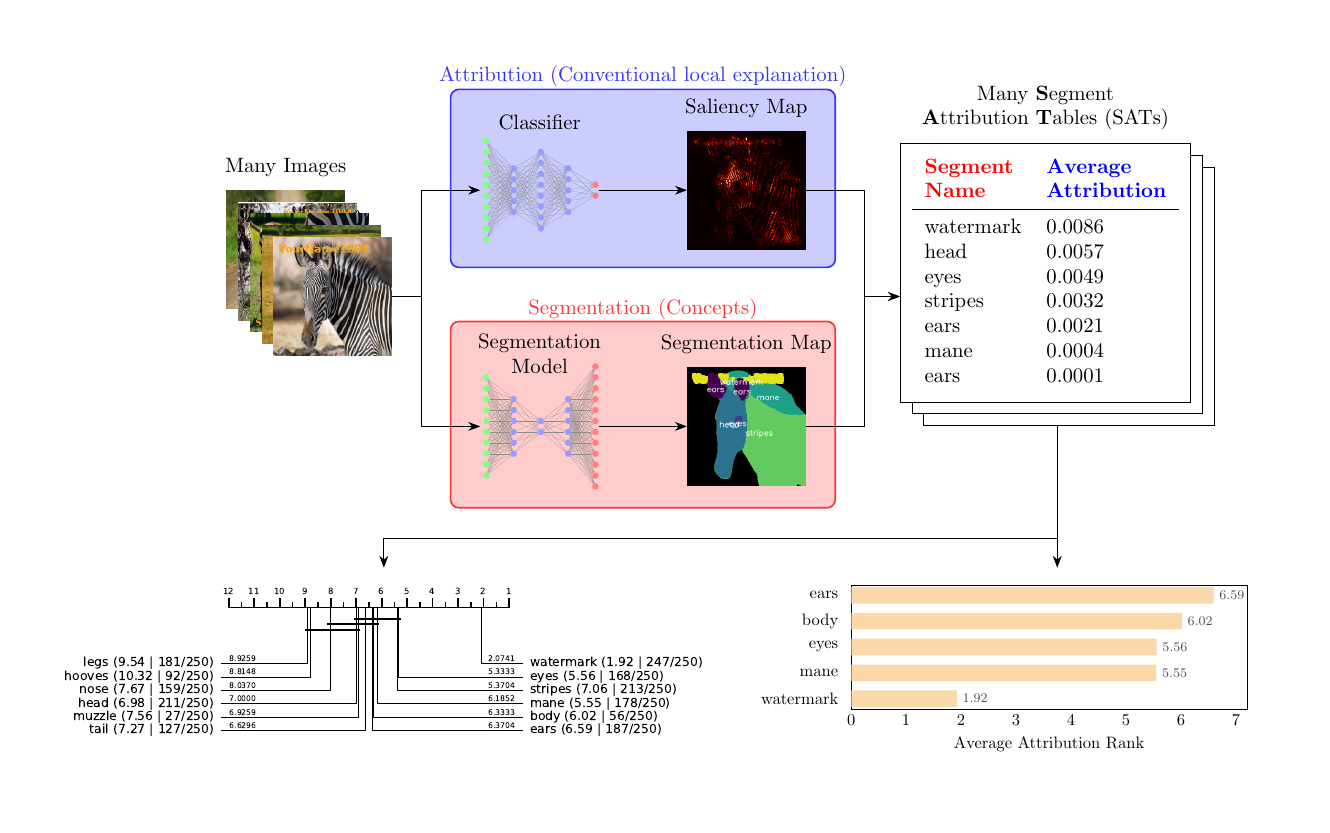}
  \vspace{-3em}
  \caption{Overview of how SATs are generated. Existing methods are used to produce
  saliency maps and segmentation maps. Each segment results in a row in the SAT, containing 
  the mean saliency value for that segment. The bottom of the figure shows two visualisations 
  constructed from aggregated SATs.}
  \label{fig:SAT_overview}
\end{figure*}

Deep learning has driven much of the recent progress in image classification~\cite{chen2021review,alom2018history}. However, as these models are increasingly deployed in critical domains such as healthcare, security, and autonomous systems, the need to understand their behaviour has grown considerably.
It is often difficult to understand why a model makes a specific
prediction, as these models rely on complex and abstract internal processes~\cite{samek2019towards}. 
Gaining insight into model behaviour is essential for diagnosing performance limitations, revealing potential biases, and increasing confidence in their reliability~\cite{molnar2025}.
Since standard evaluation metrics often overlook these dimensions, developing a deeper understanding of
model behaviour has become a key priority in machine learning research~\cite{ribeiro2016should,lapuschkin2019unmasking,lundberg2017unified}.

One particular challenge that has emerged in this context is shortcut learning. A model is said to have learned a shortcut when it focuses on spurious correlations in the data instead of learning to use features that generalise to real-world settings.
Although such strategies may lead to high performance in lab settings, they often undermine the model's ability
to generalise and act reliably in real-world scenarios.
Shortcut learning has been documented across various domains, including medical imaging, where models trained on
chest X-rays and dermoscopic skin lesion images have relied heavily on data-specific artefacts rather than clinically relevant
features~\cite{zech2018variable,degrave2021ai,winkler2019association}. 
Similarly, benchmark datasets such as ImageNet and PASCAL VOC contain spurious artefacts. Models trained on these datasets have been shown to learn shortcuts, such as relying on the presence of watermarks or counting texture patches~\cite{geirhos2020shortcut,lapuschkin2019unmasking}.
Critically, these shortcuts often vary in position across images, complicating their detection through traditional explanation methods. Furthermore, such shortcuts may occur infrequently. For example, watermarks in PASCAL VOC were observed in approximately 20\% of the `horse' training images~\cite{lapuschkin2019unmasking}. Yet even when models consistently exploit such shortcuts, explanations may not necessarily reveal this reliance in every instance~\cite{zhang2024saliency,hinns2024exposing}.

As such challenges reveal the limits of existing model assessment techniques, explanation methods become an essential
part of model evaluation as well as model understanding. Most current methods for explaining models focus on
instance-level techniques that provide local explanations for individual predictions. For example, a commonly
used local explanation method for image classification are saliency (attribution) maps, which quantify the influence of pixels or regions on the model prediction. 
While these approaches are effective at revealing what drives a specific prediction, manually reviewing large numbers of local explanations to identify recurring patterns is infeasible, particularly when such behaviours occur in only a small fraction of the data.
In contrast, direct global explanation methods attempt to summarise model activity across an entire
dataset, but they frequently fail to capture nuanced or sporadic signals, such as infrequent shortcut reliance,
that may not generalise well.

To address this gap, we introduce Segment Attribution Tables (SATs), which are produced by grounding
local saliency map explanations with contextual information, allowing for their summarisation into interpretable,
semi-global insights.
SATs summarise saliency maps based on segmented, meaningful concepts, rather than purely pixel position.
This enables the identification of recurring influential features, such as facial features in specific dog breeds or
dataset artefacts like watermarks in horse images even when these appear infrequently and at varying locations
within the dataset. \\

\noindent Our contributions are as follows:
\begin{itemize}
  \item We introduce \textit{Segment Attribution Tables} (SATs), a novel framework that summarises 
  local saliency explanations within semantically meaningful image regions to provide semi-global 
  insights into model behaviour.
  \item We empirically demonstrate, using a synthetic dataset derived from ImageNet and the BAR dataset, that SATs effectively detect shortcut features, even when these shortcuts appear in only a small fraction of instances.
  \item On the derived dataset, we further show that SATs reveal shortcut reliance that has minimal impact on accuracy and may remain hidden in standard performance evaluations.
\end{itemize}

\section{Related Work}

Local feature attribution methods, such as LIME~\cite{ribeiro2016should} and SHAP~\cite{lundberg2017unified}, 
estimate the contributions of input features to a model's predictions. While these approaches are effective in 
tabular data, their direct application to images can be less intuitive due to the absence of predefined, 
discrete features. Consequently, saliency-based methods have emerged as the dominant approach for interpreting 
deep vision models~\cite{montavon2017explaining, sundararajan2017axiomatic, bach2015pixel, 
smilkov2017smoothgrad, selvaraju2017grad}. These methods produce attribution maps highlighting influential 
image regions, typically visualised as heatmaps. 
Despite their utility, manually reviewing large numbers of saliency maps to detect subtle or infrequent behaviours is not feasible~\cite{lapuschkin2019unmasking,hinns2024exposing}.
This problem is further exacerbated by the inherent noise and instability common to saliency methods~\cite{zhang2024saliency}. 
Explain Any Concept (EAC)~\cite{sun2023explain}, in contrast, focuses solely on providing local explanations 
by combining segmentation from the Segment Anything Model (SAM) with Shapley values computed via a 
local surrogate model to address computational demands. Although EAC demonstrates quantitatively 
superior local explanations compared to other methods and user preference, it does not inherently 
support aggregation since SAM, by default, does not produce segment labels.

\begin{figure*}[t]
    \centering
    \includegraphics[width=\linewidth]{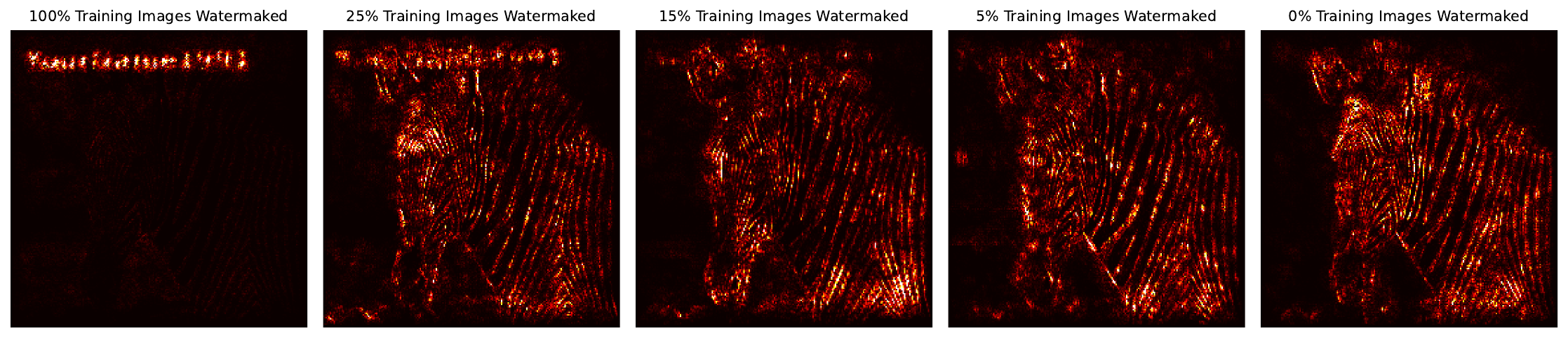}
    \caption{LRP-Z Salinecy maps for models trained with varying amounts of watermarked training images.}
    \label{fig:watermark_vary_heatmap}
\end{figure*}

At the opposite end of the spectrum, global explanation methods aim to provide high-level summaries of model 
behaviour across a whole dataset. Techniques such as Testing with Concept Activation Vectors (TCAV)~\cite{kim2018interpretability} and Concept Bottleneck Models (CBMs)~\cite{koh2020concept} connect model 
predictions to human-interpretable concepts~\cite{ghorbani2019towards}. TCAV measures the sensitivity of 
model outputs to predefined concepts via activation vectors~\cite{kim2018interpretability}, while CBMs 
explicitly incorporate concept prediction as an intermediate classification step~\cite{koh2020concept}. 
Despite their conceptual appeal, methods based on concepts typically require specialised datasets or 
model retraining~\cite{yuksekgonul2022post}, exhibit fragility~\cite{li2025interpretability}, 
and may produce concepts that do not align well with human intuition~\cite{yeh2020completeness}.

To bridge the gap between detailed local explanations and broad global summaries, several approaches aggregate 
or structure local outputs to facilitate more practical model analysis. For instance, Counterfactual 
Frequency (CoF) Tables~\cite{hinns2024exposing} generate counterfactuals, identifying segments whose 
alteration changes predictions, which can be aggregated based on the name of the segment edited. 
CoF tables also suffer from coverage limitations common to counterfactual methods, where no explanation is provided for some instances, and computational inefficiencies from repeatedly classifying edited images. 
Additionally, the image edits used must be carefully designed, as simplistic 
modifications like blurring often perform poorly with modern large image models. 
CoF tables explicitly address the limitation in EAC of SAM not providing labels by using 
segmentation models that provide semantic labels, such as Grounded SAM~\cite{ren2024grounded}.
Spectral Relevance Analysis (SpRAy)~\cite{lapuschkin2019unmasking} clusters Layer-wise Relevance Propagation (LRP) saliency maps to reveal common decision-making patterns; however, this means it prioritises the position of influential areas, rather than their semantic meaning. Saliency maps themselves indicate only which pixels are influential, without providing any contextual understanding of what those pixels represent. Consequently, clustering these maps tends to group explanations primarily by spatial patterns, causing positional information to dominate over the underlying semantic concepts. For example, if a watermark shortcut appeared in different positions across images, SpRAy would likely return multiple separate clusters, each corresponding to a different position, rather than identifying a single watermark concept influencing the model. \\
Like our approach, Global Saliency~\cite{pfau2019global} quantifies model biases by averaging attributions over predefined semantic segments. 
It also calculates average attributions for segments, but focuses exclusively on a single predetermined segment identified by a specifically trained segmentation model.
Although this method is effective for identifying whether models rely on known features, such as ink markings in skin lesion diagnosis, its interpretability is limited without comparative reference points. The global saliency value depends entirely upon the chosen saliency method, dataset, and model, meaning that an isolated numeric score is difficult to interpret. For instance, a global saliency of 15 for a segment is not meaningful on its own. The value becomes more interpretable when compared with something else. For example, it is clearer to state that a model gives higher saliency to a segment than a different version of the model, or that ink markings are assigned higher saliency than the lesion itself.
Additionally, it cannot identify previously unknown influential patterns, relying entirely on 
prior knowledge of the segment to be analysed.

\section{Method}
\label{sec:method}

The process of producing and aggregating SATs involves several considerations, which we address in this section.
Figure~\ref{fig:SAT_overview} provides an overview of the process. 
We exclude details of generating saliency or segmentation maps, as these are thoroughly covered by prior works. 
We emphasise that our approach is compatible with any form of heatmap, including attention 
maps, SHAP, and LIME.

\subsection{Producing SATs}
Let \(I \in \mathbb{R}^{h \times w}\) denote an image represented as a matrix of pixels of height $h$ 
and width $w$. The number of channels (e.g., single-channel greyscale or multi-channel RGB) is 
irrelevant to this definition.
Similarly let \(A \in \mathbb{R}^{h \times w}\) be the corresponding saliency (attribution) map, 
with \(A_{x,y}\) indicating the saliency at pixel \((x,y)\). Let \(S = \{ (k_1, M_1), (k_2, M_2), \dots, (k_n, M_n)\}\) be a set of \(n\) segments, each defined 
by a tuple comprising a segment name $k_i$ and a boolean mask \(M_i \in \{0,1\}^{h \times w}\).
Here, $M_i$ is the boolean mask for segment \(i\), where \((M_i)_{x,y} = 1\) indicates that 
pixel \((x,y)\) belongs to segment \(i\).
Pixels may appear in multiple segment masks by this definition.
Optionally, we could instead require that the masks be pairwise disjoint,  
i.e.\ \(\forall\,i \neq j,\,(M_i \odot M_j)_{x,y}=0\) for all \(x,y\), although we do not impose this constraint here.
We refer to the set \(S\) as the segmentation map.

We adopt the definition of mean saliency $\bar a_i$ from~\cite{pfau2019global}, modified to suit our 
multiple-segment context:

\[
\bar a_i
= \frac{1}{Z_i}\sum_{x,y}(M_i \odot A)_{x,y},
\quad
Z_i = \sum_{x,y}(M_i)_{x,y}
\]

\noindent
where $Z_i$ is the number of pixels in segment $i$, 
$1 \le x \le w$, 
$1 \le y \le h$, 
and $\odot$ is the Hadamard (element-wise) product.

To accommodate methods that produce negative attributions, we first compute the mean saliency over each 
segment and then take its absolute value, denoted $\bigl|\bar a_i\bigr|$.
Applying the absolute value after averaging causes opposing signals within a segment to cancel out. 
As a result, segments with a strong, coherent consensus toward one decision rank higher than those with 
mixed, noisy signs.
We further define the local rank of the mean attribution within a SAT, with rank 1 corresponding to the 
segment with the highest mean attribution. This facilitates fairer comparisons across SATs and reduces 
the need for explicit normalisation of saliency maps.

In addition to mean attribution and segment name, we enrich SATs with contextual metadata, 
including mask size, segment position (e.g., top-left, centre-centre, bottom-right), and total attribution. 
Practically, we also add segment and images IDs, as well as the saliency method used, 
given that we compute multiple SATs from different methods for the same image and segmentation map.

\subsection{Aggregating SATs}
SATs can be aggregated by simple concatenation and averaging, but because different images often yield different segment sets, the resulting tables may display non-identical segment names and varying entry counts. We employ straightforward strategies to address this heterogeneity.

For mean attribution, we define two aggregation methods: \textbf{Relative aggregation}, which calculates the mean attribution of each segment name across all SATs irrespective of its frequency; and \textbf{Absolute aggregation}, which first compiles the complete set of segment names \(K^*\) and then appends any names in \(K^*\setminus K\) to each SAT with an attribution value of zero. In rank-based aggregations, these supplementary segments are assigned a rank one greater than the SAT’s original number of segments.

\begin{table}[h]
    \centering
    \caption{Relative aggregate SAT constructed by averaging each segment’s mean attribution across all images in the controlled watermark setting, as shown in Figure~\ref{fig:SAT_overview} and discussed in the results section.}
    \label{tab:ex_asat}
    \vspace{0.5em}
    \begin{tabular}{ll}
        \toprule
        Segment Name (\(k_i\)) & Aggregate Mean Attribution (\(\overline{\bar{a}}_i\)) \\
        \midrule
        watermark & 0.0135 \\
        mane      & 0.0051 \\
        \(\vdots\) & \(\vdots\) \\
        hooves    & 0.0013 \\
        \bottomrule
    \end{tabular}
\end{table}

Once aggregated, SATs can be explored through many different means, such as grouping according to any 
of the information present, like segment name and position, or filtering based on mask size. 
The aggregated tables (as shown in Table~\ref{tab:ex_asat}) can be summarised clearly through tables or visualisations, 
examples of which are provided in the results section.

\section{Results}
\label{sec:results}
\FloatBarrier

In this section, we show insights that can be gained from SATs across 
a variety of datasets, classifiers, segmentation methods and saliency methods.
For all our experiments, we produce saliency maps using the iNNvestigate package~\cite{JMLR:v20:18-540}.

\subsection{Understanding Model Behaviour}

To show how SATs can be used to understand model behaviour, we first demonstrate 
their use explaining a well performing pre-trained model. For this we use the VGG16 classifer~\cite{simonyan2014very}, 
using tensorflows pre-trained weights on ImageNet-21k~\cite{russakovsky2015ImageNet}.

We use the Deep Taylor decomposition method~\cite{montavon2017explaining} to produce the saliency maps for
this experiment.
The segmentation maps are made using the DinoX segmentation model~\cite{ren2024dinoxunifiedvisionmodel},
which is an open-set segmentation model, meaning it can segment any object in the image, not just a fixed set of classes.
In order to control the segmentation fidelity and maintain a consistent number of segments, we prompt the model to divide the images into twelve segments chosen to be characteristic of each class. These segments were selected by asking ChatGPT to propose twelve regions likely to be indicative of the given class.
we then edited these lists to ensure that the segments would be detectable and not too abstract.
If no segments are prompted, the model would decide the segments itself, based on confidence thresholds
passed to it.
We also pad the masks by two pixels, as the saliency maps tend to assign the highest importance to segment edges, and the masks are often conservative.

We show visualisations of SATs for three classes: Chihuahua, Firetruck and Zebra. 
For each class we choose one hundred images, discarding those that are mislabelled, overshadowed by another subject, or ambiguous and difficult to classify even with prior knowledge (for example, where both a Chihuahua and a feather boa appear, and either could be a valid prediction). These issues in ImageNet are already well documented \cite{northcutt2021labelerrors}.

To visualise the SATs, we use critical difference diagrams, implemented by the Aeon package~\cite{aeon24jmlr}. 
These diagrams show the average rank of mean attribution for each segment,
with solid bars connecting segments that are not statistically significantly different from each other.
Statistical significance is assessed using Wilcoxon signed-rank tests, with groups of segments formed using the Holm correction.

\begin{figure}[]
    \centering

    \begin{subfigure}{\linewidth}
        \centering
        \includegraphics[width=\linewidth]{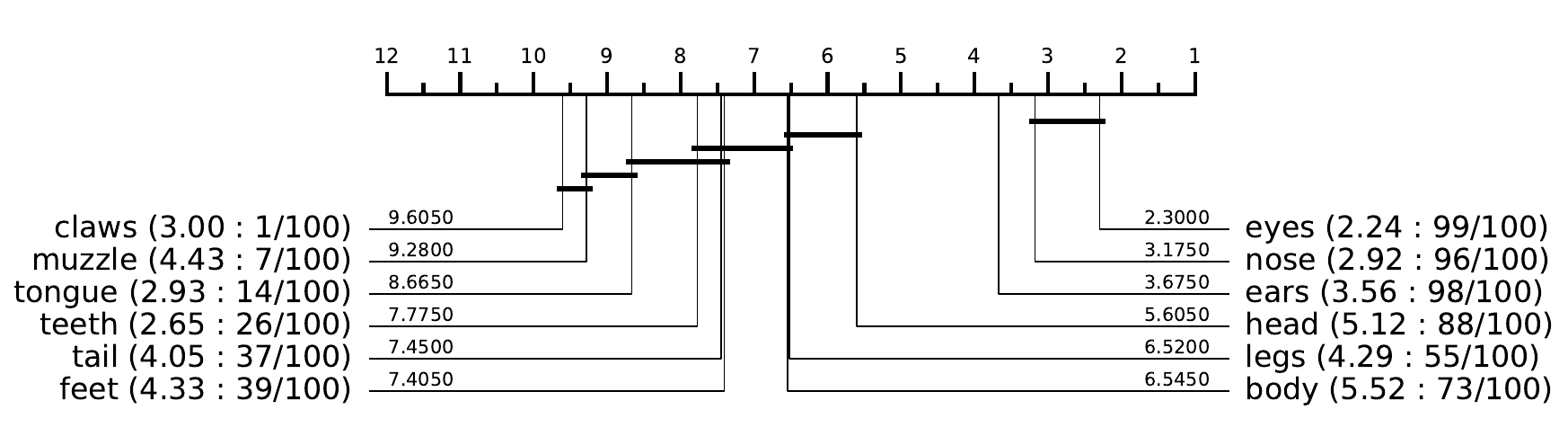}
        \caption{Chihuahua}
        \label{fig:deep_taylor_chihuahua}
    \end{subfigure}

    \begin{subfigure}{\linewidth}
        \centering
        \includegraphics[width=\linewidth]{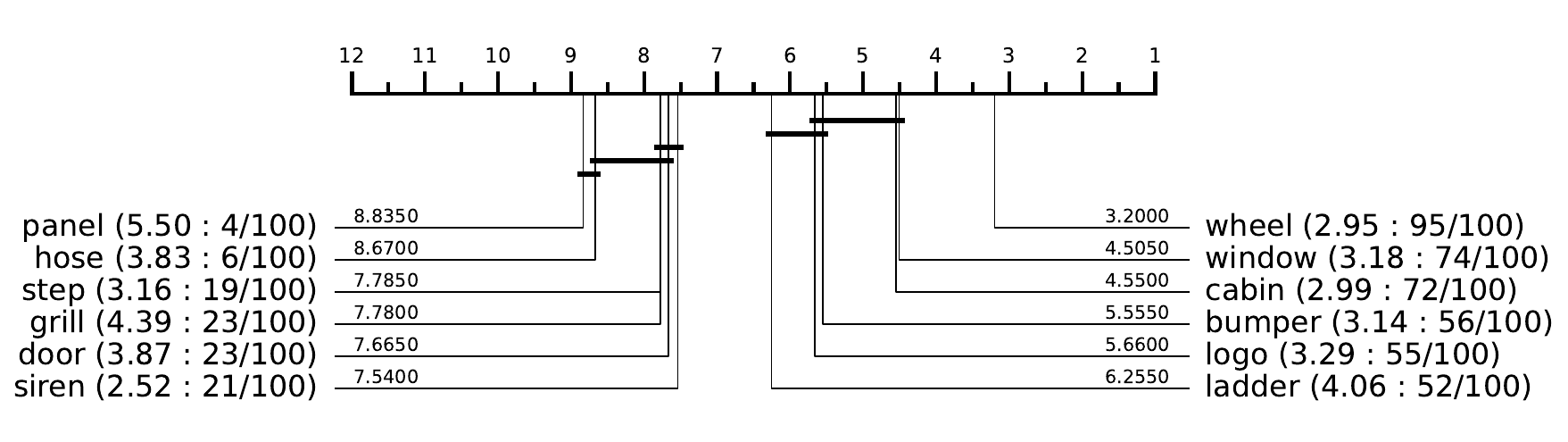}
        \caption{Firetruck}
        \label{fig:deep_taylor_firetruck}
    \end{subfigure}

    \begin{subfigure}{\linewidth}
        \centering
        \includegraphics[width=\linewidth]{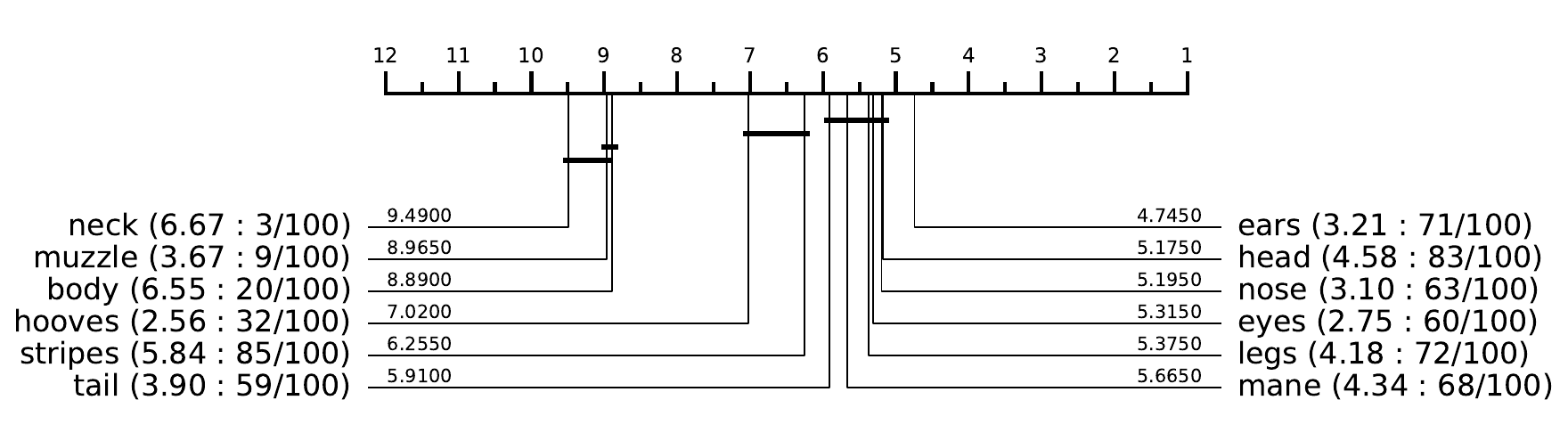}
        \caption{Zebra}
        \label{fig:deep_taylor_zebra}
    \end{subfigure}
    
    \caption{Critical difference diagrams of SATs using Deep Taylor for the Chihuahua, Firetruck, and Zebra classes in ImageNet.}
    \label{fig:deep_taylor_comparison}
\end{figure}

In Figure~\ref{fig:deep_taylor_chihuahua}, we show a critical difference diagram for the 100 
images of the Chihuahua class. You can see that the most impactful features are all related to the
dog's face, with the eyes being the most important segment, and that there isn't a statistically
significant difference between the ears and head. This is quite logical behaviour to classify 
a chihuahua out of the 21,841 classes in ImageNet-21k, especially when it is considered that the 
photos in ImageNet largely show the dog towards the camera.


In Figure~\ref{fig:deep_taylor_comparison}, we show the critical difference diagrams for all three
classes: Chihuahua, Firetruck and Zebra.

We compare six different saliency methods on the Chihuahua class and observe a large amount of agreement between them. However, this agreement can be misleading, as the critical difference diagrams are based on the complete aggregation, which ranks undetected segments last by default, as explained in the method section.
As a result, the critical difference diagram is highly dependent on the segmentation process, and segments that appear more frequently across SATs are likely to be ranked higher.
Although this provides a clear overview of which segments are both detectable and important across the images while preserving the paired significance test, it can obscure rare edge cases.
To mitigate this effect, we also display the relative aggregation mean rank alongside the number of images in which each segment appears (shown in brackets after the segment name) in all diagrams.

To gain this form of insights through saliency maps alone, a human would need to manually review
all images for each class. Here, we demonstrate 100 images to represent the whole class, but 
in practice, many more images could be used to give a more accurate overview. This overview 
could then be explored, for example, filtering out all images that do not detect body for 
the Chihuahua class.

\subsection{Shortcuts}

\subsubsection{Biased Action Recognition}

To first demonstrate the potential to identify shortcuts using SATs, we use the Biased Action Recognition 
(BAR) dataset~\cite{nam2020learning}. BAR has six classes consisting of different actions biased towards 
specific environments, shown in Table~\ref{tab:BAR_info}.
\begin{table}[h]
\centering
\caption{Details of the BAR dataset \cite{nam2020learning}.}
\begin{tabular}{llrr}
\toprule
\textbf{Action Class} & \textbf{Environment Bias} & \textbf{Training} & \textbf{Test} \\
\midrule
Climbing   & Rock Wall      & 326 & 105 \\
Diving     & Underwater     & 520 & 159 \\
Fishing    & Water Surface  & 163 &  42 \\
Racing     & Road           & 336 & 132 \\
Throwing   & Playing Field  & 317 &  85 \\
Vaulting   & Sky            & 279 & 131 \\
\bottomrule
\end{tabular}
\label{tab:BAR_info}
\end{table}

This bias only exists in the training set, with the testing set showing actions being in 
different environments. This allows us to assess the extent to which the model has learned a shortcut based on the environment, by comparing its accuracy on the test set with that on the training set.

We aim to use SATs to understand the model’s reliance on shortcuts without requiring access to an OOD test set for validation, as this is often not realistic in practice.
To do this, we train a simple CNN on the training set, and create SATs for each instance of a number of classes.
We train a second CNN, with the same architecture, but mixing a selection of images from the test set 
into the training set and perform some data augmentation such as random cropping and flipping, to
produce a less-biased model.

\begin{figure*}[!t]
    \includegraphics[width=\linewidth]{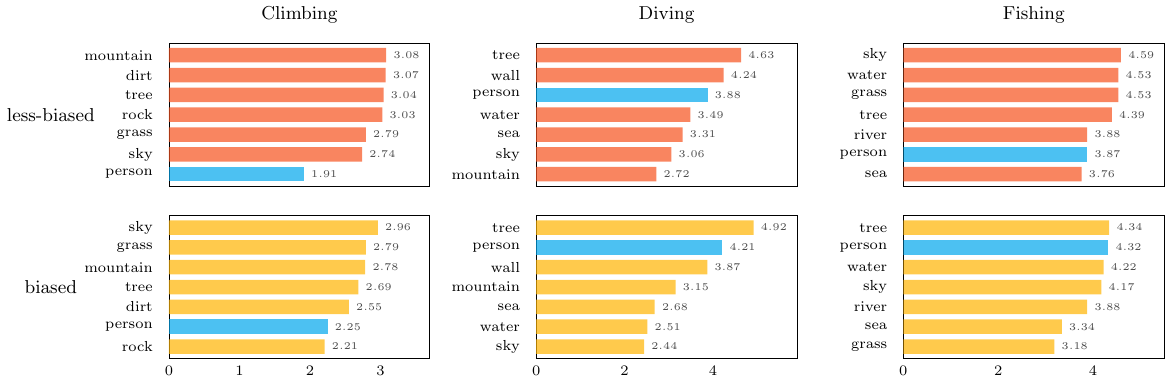}
    \caption{The seven most influential segments (Y-axis) for the BAR dataset, based on the relative aggregate mean attribution (X-axis), shown for three selected classes (Climbing, Diving, Fishing) and two models (biased and less-biased).}
    \label{fig:BAR_matrix}
\end{figure*}

We show a comparison of these two models in Figure~\ref{fig:BAR_matrix}, where we show the seven most 
influential segments for three classes; climbing, diving and fishing.
For all three classes, one might expect the most influential segment to be the person, 
as a person is performing the action, however, we see this for only one of the six aggregate 
SAT plots we show.

\FloatBarrier   

Figure~\ref{fig:BAR_matrix} shows that the less-biased model has learned to place a higher importance on the person
than the biased model across all three classes. We also see that the importance of the shortcuts segments,
such as rock, water and sea generally decreases in the less-biased model. 
From these plots, we still see unwanted behaviour in the less-biased model, with the relatively low importance
of the person segment for the diving and fishing classes, as well as strong reliance still on background 
segments. 
This is expected, as the less-biased training data still contains a substantial number of instances in which the environment shortcut is present.

\subsubsection{Toy Watermark Dataset}

To demonstrate the ability of SATs to highlight behaviours that only affect a small
proportion of the dataset, we create a toy dataset.
We use horse and zebra classes from ImageNet, and place watermarks in a random corner of the zebra images.
The watermarks vary in text and are placed in one of the four corners with a slight jitter, while maintaining a consistent colour and size.

The dataset is split into training and testing sets. 
To directly compare model behaviour, two versions of the test images were created: one without watermarks (clean) and one fully watermarked (watermarked). Accuracy was then measured against both the clean and watermarked test sets. 
Finally, multiple training datasets were created by systematically varying the proportion of watermarked images. 
This variation weakens the shortcut's effectiveness, allowing us to test how sensitively SATs can detect the model's reliance on it.

\begin{figure}
    \centering
    \includegraphics[width=0.48\textwidth]{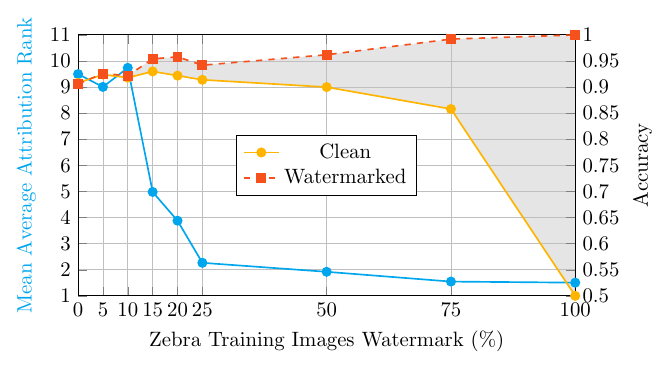}
    \caption{Average attribution rank of the watermark segment (calculated on the watermarked test set) and accuracy (reported on both clean and watermarked test sets), for models trained with varying prevalence of watermark shortcut.}
    \label{fig:watermark_acc_rank}
\end{figure}

For each training set, we train a CNN and produce saliency maps using the LRP-Z for the watermarked test set. 
Once more we segment the images using DinoX, and choose the same 12 labels as in the 
ImageNet experiment, but replace the least impactful segment (neck) with watermark.
We produce SATs for the models on the watermarked test set.

\begin{figure*}[!t]
    \centering
    \includegraphics[width=\textwidth]{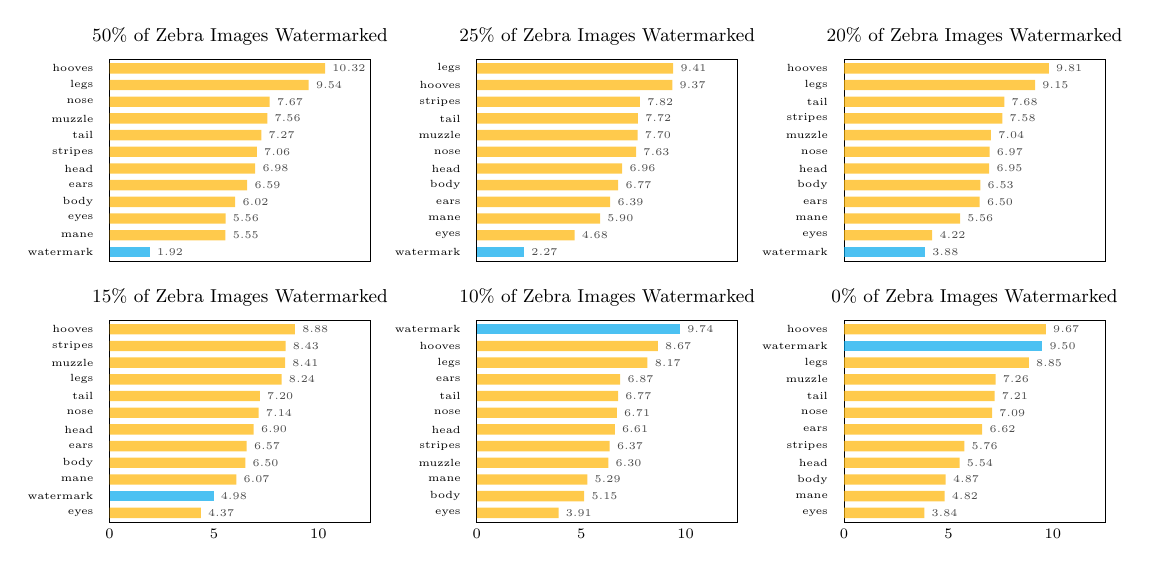}
    \caption{Relative aggregate mean attribution rank (X-axis) for the 12 chosen segments (Y-axis), computed using the LRP-Z saliency method. Subplots show visualisations of SATs for models trained with different proportions of watermarked images. Results are produced on the 100\% watermarked test set. Watermark segment is highlighted in blue.}
    \label{fig:watermark_group_plot}
\end{figure*}

In Figure~\ref{fig:watermark_group_plot}, we present visualisations of the SATs for the different models. When 15\% or more of the zebra images are watermarked, the SATs clearly reveal a strong influence of the watermark on model behaviour. Notably, the consistently high ranking of segments such as eyes across models indicates that, even when the watermark shortcut is present, the models continue to learn meaningful features for classification. This outcome reflects the data distribution. Even when 50\% of zebra images are watermarked, 75\% of all images (including horse and zebra classes) remain unmarked, meaning the model still benefits from learning genuine distinguishing features.

Figure~\ref{fig:watermark_acc_rank} shows the mean rank of average attribution and the change in performance from the clean to the watermarked test set. We observe that test accuracy remains relatively stable until after 25\% of zebra images are watermarked. However, SATs begin to highlight the watermark shortcut much earlier, with a clear signal emerging from around 15\% watermarking. This reflects the earlier point that the model continues to learn the true classification task, making use of meaningful segments such as eyes, while also learning the shortcut in parallel. It is only as the shortcut becomes increasingly predictive that the model begins to rely on it more heavily.

After the 25\% watermarking threshold, the attribution rank begins to plateau, from approximately 2.3 to 1.5, while the change in accuracy grows markedly, from 3.6\% to 50\%. 
A 50\% change in accuracy indicates that the model has learned only the watermark shortcut, predicting images as zebra when the watermark is present and as horse when it is absent.
This results in 50\% accuracy on the clean set, where no watermarks are present and all images are incorrectly predicted as horse, and perfect accuracy on the watermarked set, where the shortcut alone enables correct predictions.

This demonstrates that SATs can reveal subtle model behaviours that are not captured through conventional accuracy metrics alone. Detecting such behaviours is of particular importance, as real-world shortcuts appearing in the majority of images would be simpler to detect through conventional means. For instance, \cite{lapuschkin2019unmasking} highlights a watermark shortcut in the Pascal VOC dataset that appears in approximately 20\% of images.

\section{Conclusion and Future Work}

We introduce \textit{Segment Attribution Tables} (SATs) for summarising saliency explanations to uncover semi-global patterns. By aggregating local explanations within semantic segments, SATs reveal consistent influences of meaningful image regions across large datasets.
This systematic approach addresses the infeasibility of manually reviewing numerous individual explanations, especially when features are subtle or occur infrequently. Infrequent patterns require extensive review to detect, while subtle ones may be overshadowed by more obvious, frequent behaviours. SATs effectively highlight influential segments and expose spurious correlations, such as background artefacts or dataset-specific watermarks. This allows for comprehensive evaluations of models and represents a meaningful advance towards ensuring critical properties of AI systems, including reliability, fairness, and trustworthiness. By enabling deeper insights into model predictions, SATs support the development and deployment of trustworthy models that generalise effectively to real-world applications.

The current work explores only a limited range of aggregation techniques for SATs, and future research could examine additional methods, like those proposed by~\cite{ribeiro2016should} or~\cite{van2019global}, to assess their effects on aggregate SATs. Our visualisations are currently limited to bar charts and critical difference diagrams, so expanding visualisation methods could enhance interpretability. Furthermore, the suitability of SATs depends significantly on the segmentation maps used. Issues such as incorrect labels, undetected objects, or inappropriate granularity (e.g., identifying a whole object rather than its parts) can reduce effectiveness. While targeted prompting with open-set models mitigates some of these problems, this method requires expert oversight. Future advances in segmentation and vision–language models may help address these limitations. Finally, future research could explore aggregating multiple local explanation methods to yield more robust explanations, as suggested by~\cite{rieger2019aggregating}. Investigating how SATs might guide model optimisation, either by prioritising key segments, such as ensuring that person ranks higher in importance than rock, or by minimising the influence of unwanted segments, such as watermarks, presents a promising direction for future work.

\section*{Acknowledgments}
We acknowledge the support of the Onderzoeksprogramma Artifici\"ele Intelligentie (AI) Vlaanderen (FAIR). We are also grateful to IDEA Research for providing credits for their Dino-X model.

\bibliography{ref}

\section{Appendix}
\begin{figure*}[ht]
    \centering

    \begin{subfigure}{0.7\linewidth}
        \centering
        \includegraphics[width=\linewidth]{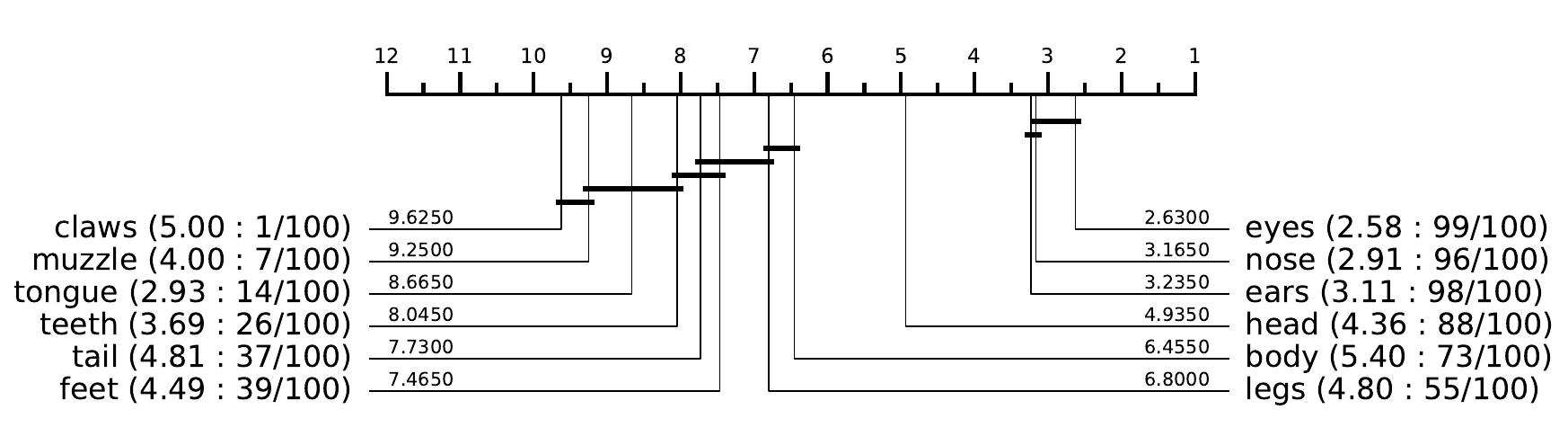}
        \vspace{-2.4em}
        \caption{Deep Taylor}
    \end{subfigure}

    \begin{subfigure}{0.7\linewidth}
        \centering
        \includegraphics[width=\linewidth]{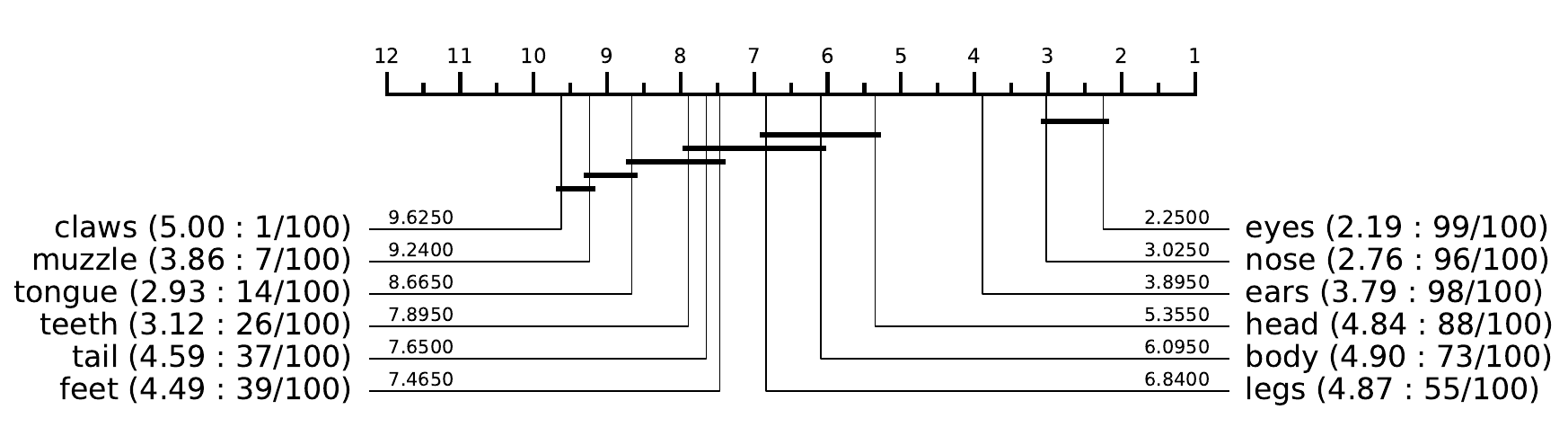}
        \vspace{-2.4em}
        \caption{LRP-$\epsilon$}
    \end{subfigure}

    \begin{subfigure}{0.7\linewidth}
        \centering
        \includegraphics[width=\linewidth]{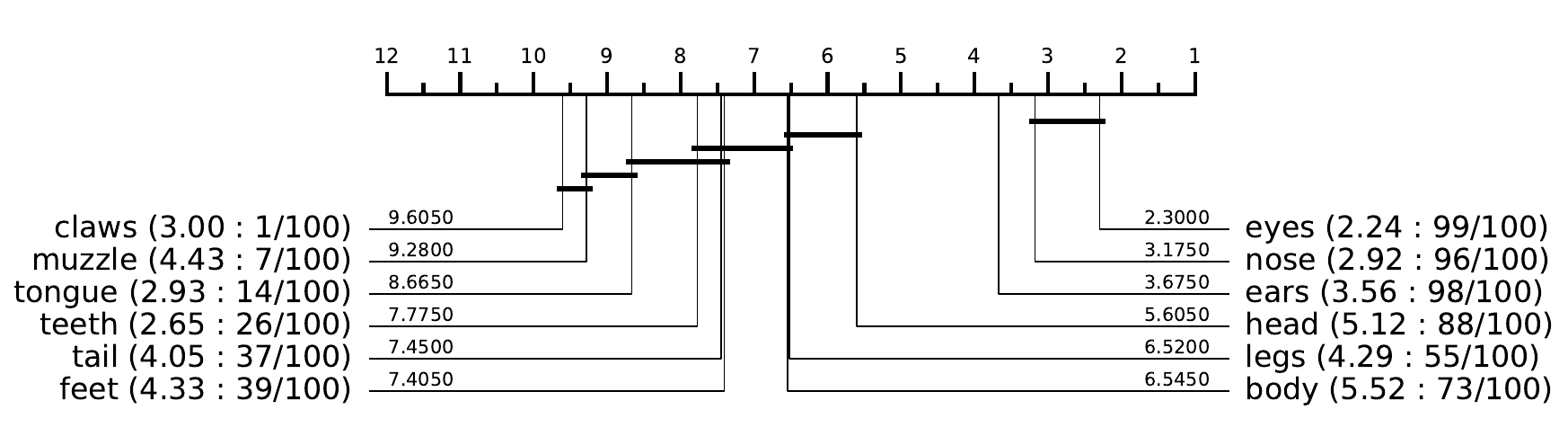}
        \vspace{-2.4em}
        \caption{LRP-Z}
    \end{subfigure}

    \begin{subfigure}{0.7\linewidth}
        \centering
        \includegraphics[width=\linewidth]{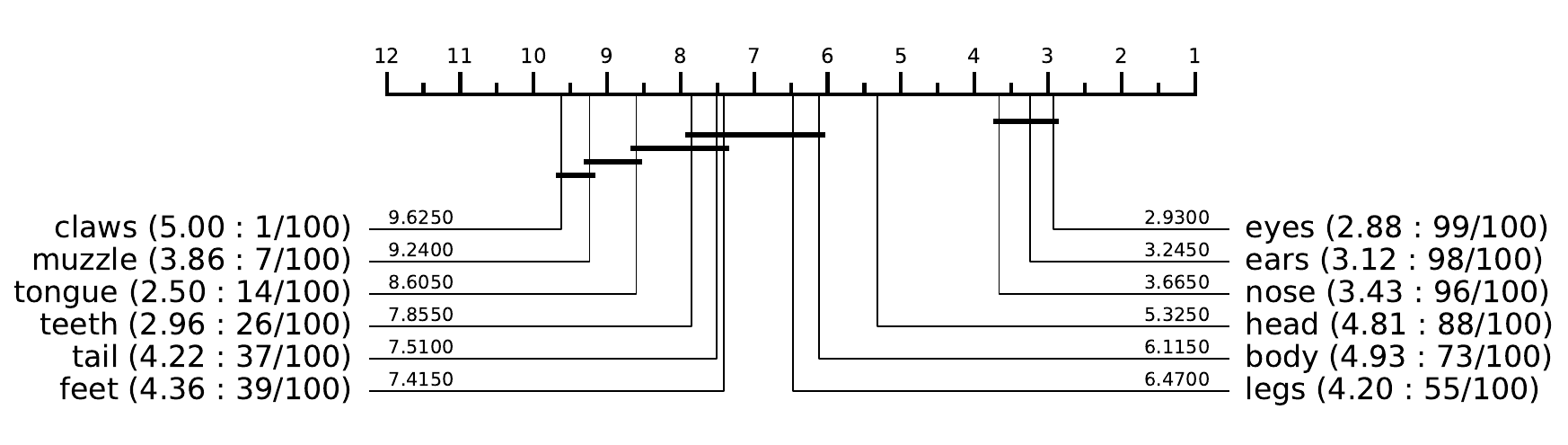}
        \vspace{-2.4em}
        \caption{Integrated Gradients}
    \end{subfigure}

    \begin{subfigure}{0.7\linewidth}
        \centering
        \includegraphics[width=\linewidth]{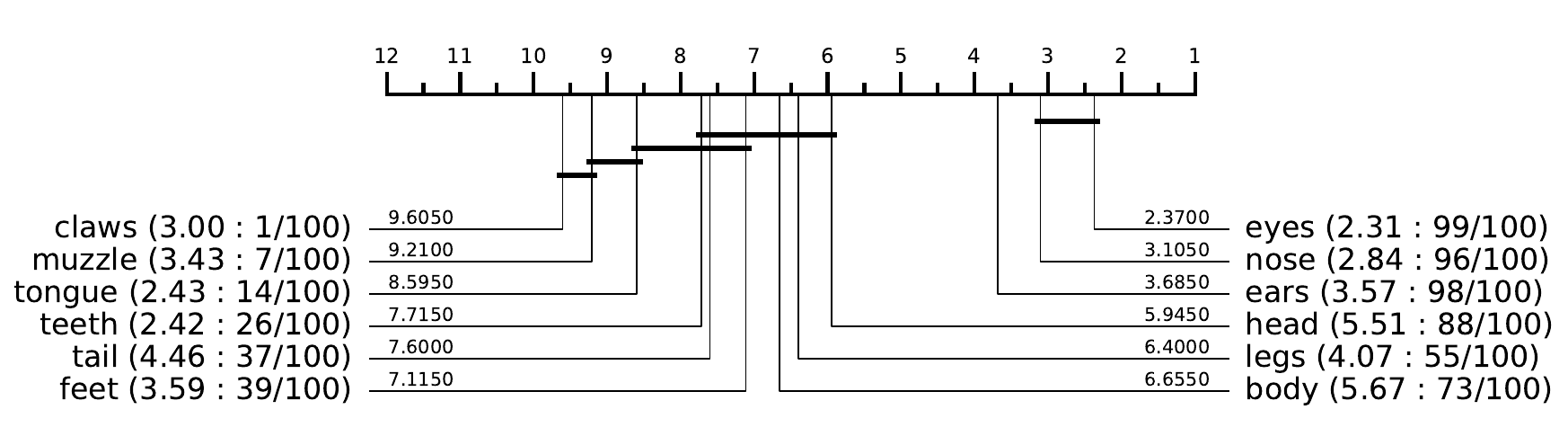}
        \vspace{-2.4em}
        \caption{Gradient}
    \end{subfigure}

    \begin{subfigure}{0.7\linewidth}
        \centering
        \includegraphics[width=\linewidth]{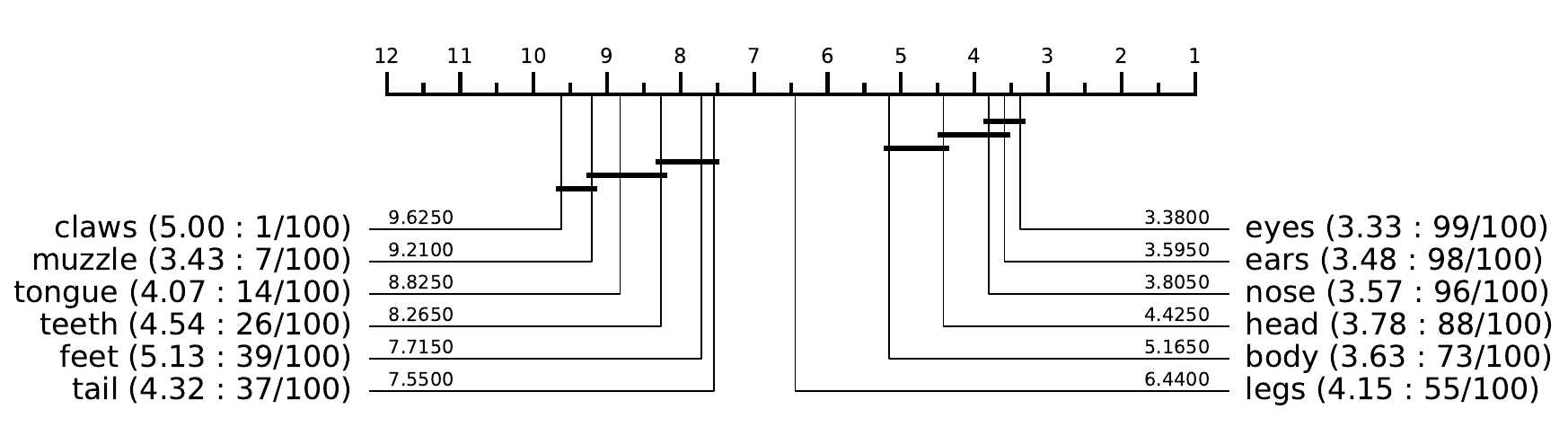}
        \vspace{-2.4em}
        \caption{SmoothGrad}
    \end{subfigure}

    \caption{Critical difference diagrams for different saliency methods for the ImageNet \textit{Chihuahua} class. The diagrams display absolute aggregation values by default, which are also used for computing statistical significance. Labels additionally include the relative aggregation and the number of unique images in which each segment appears.}
    \label{fig:critical_difference_matrix}
\end{figure*}

\end{document}